\documentclass[lettersize,journal]{IEEEtran}
\usepackage{amsmath,amsfonts}
\usepackage{algorithm}
\usepackage{array}
\usepackage[caption=false,font=normalsize,labelfont=sf,textfont=sf]{subfig}
\usepackage{textcomp}
\usepackage{stfloats}
\usepackage{url}
\usepackage{verbatim}
\usepackage{graphicx}
\usepackage{cite}
\usepackage{cite}
\usepackage{amsmath,amssymb,amsfonts}
\usepackage{graphicx}
\usepackage{textcomp}
\usepackage{xcolor}
\usepackage{setspace}
\usepackage{algorithm}
\usepackage{algpseudocode}
\usepackage{booktabs} 
\usepackage{multirow} 
\usepackage{array}
\usepackage[table]{xcolor}

\def\BibTeX{{\rm B\kern-.05em{\sc i\kern-.025em b}\kern-.08em
    T\kern-.1667em\lower.7ex\hbox{E}\kern-.125emX}}
\begin{document}

\title{FAST: A Framework for Aligned Sampling and Training in Parallel Reinforcement Learning for Autonomous Driving}

\author{Bonan~Wang$^{*}$,
        Letian~Tao$^{*}$,
        Bin~Shuai,
        Jiaxin~Gao, 
        Wenxin~Zhao,
        Wei~Xiong,\\
        Kehua~Sheng, 
        Bo~Zhang,
        Yang~Guan$^{\dag}$,
        and Shengbo~Eben Li$^{\dag}$ ~\IEEEmembership{Senior Member,~IEEE}
\thanks{*\,Authors contributed equally; \dag\,Corresponding author.}
\thanks{Bonan Wang, Letian Tao, Bin Shuai, Jiaxin Gao, Wenxin Zhao and Yang Guan are with the School of Vehicle and Mobility, Tsinghua University, Beijing 100084, China. }
\thanks{Wei Xiong, Kehua Sheng and Bo Zhang are with Didi Voyager Labs, DiDi Autonomous Driving. }
\thanks{Shengbo Eben Li are with the School of Vehicle and Mobility and the College of AI, Tsinghua University, Beijing 100084, China. }
\thanks{E-mail: {lishbo@tsinghua.edu.cn, yguan@tsinghua.edu.cn}}
\thanks{This work is supported by the National Natural Science Foundation of China with 92582205 and the Key Program of the Beijing Municipal Natural Science Foundation with L257002.}
\thanks{This work has been submitted to the IEEE for possible publication. Copyright may be transferred without notice, after which this version may no longer be accessible.}
}

\markboth{Journal of \LaTeX\ Class Files,~Vol.~14, No.~8, August~2021}%
{Shell \MakeLowercase{\textit{et al.}}: A Sample Article Using IEEEtran.cls for IEEE Journals}


\maketitle

\begin{abstract}
Deep reinforcement learning is pivotal for closed-loop autonomous driving yet remains constrained by severe bottlenecks in sampling efficiency. Standard parallel sampling mitigates this but suffers from the straggler effect, where the premature termination of a single environment necessitates a synchronized batch re-initialization, leading to suboptimal sample utilization and prohibitive re-initialization latency. To address this, we propose FAST, a synchronous parallel framework tailored for closed loop simulation. Specifically, FAST employs Dynamic Parallel Sampling Alignment (DPSA) to maintain vectorization synchronization by extending terminated episodes via virtual continuation, thereby decoupling the sampling loop from individual terminations. By dynamically triggering global truncation based on the termination rate of parallel clips, FAST effectively eliminates the bottleneck of premature resets without sacrificing data diversity. Furthermore, to strictly preserve theoretical consistency, we incorporate a Scaled Mask Padding Optimization (SMPO) that leverages validity masking and adaptive loss normalization to nullify the bias from auxiliary padding data. Empirical evaluations demonstrate that FAST achieves at least a 1.78$\times$ wall-clock speedup over the single-clip baseline while preserving statistical unbiasedness.
\end{abstract}

\begin{IEEEkeywords}
Autonomous Driving, Deep Reinforcement Learning, Parallel Sampling, Sampling Efficiency
\end{IEEEkeywords}

\section{Introduction}

\IEEEPARstart{D}{eep} Reinforcement Learning (DRL) has emerged as a fundamental paradigm in autonomous driving (AD), enabling robust decision-making within high-dimensional and non-stationary environments \cite{li2023reinforcement,hu2026long,huang2025vlm,mnih2015human}. Integrating high-dimensional sensory processing with decision-making allows DRL agents to master critical competencies such as path planning, trajectory optimization, and collision avoidance through continuous trial and error. Achieving robust policy performance thus requires training on massive interaction data, often demanding millions or billions of simulation steps to cover long-tail corner cases and ensure safety generalization \cite{liao2026addressing,wang2025beyond,zhang2025carplanner}.

Driven by this immense data demand, the continuous ingestion of high-quality interaction data represents a critical component within the pipeline of DRL \cite{huang2022distributed,guan2020centralized}. In contrast to open-loop paradigms that leverage static, pre-recorded datasets for model optimization, closed-loop simulation necessitates the dynamic generation of environmental feedback in response to the agent's actions \cite{codevilla2018end}. In high-fidelity autonomous driving scenarios, the interactive sampling process encounters a severe efficiency bottleneck arising from the high dimensionality of sensory data \cite{dosovitskiy2017carla}. The constrained transmission of these massive data volumes across hardware hierarchies significantly hampers sampling efficiency, thereby limiting the aggregate training throughput \cite{li2022metadrive,espeholt2019seed}.

To bypass the strict temporal dependencies of data acquisition, certain  paradigms leverage off-policy learning as a potential solution
\cite{espeholt2018impala,petrenko2020sample}. While off-policy methods can alleviate synchronization overhead by utilizing asynchronous data, they introduce significant instability due to the divergence between behavioral and target distributions. Such misalignments frequently lead to value overestimation and extrapolation errors, which are detrimental to safety-critical tasks \cite{fujimoto2019off, farid2026policy}. Consequently, on-policy learning remains the preferred choice for autonomous driving due to its superior stability and rigorous distribution alignment \cite{guan2021direct}. Although parallelization offers a viable pathway to scale these on-policy methods, the deployment of distributed architectures is encumbered by a fundamental tension between sampling throughput and distributional alignment. Specifically, because these closed-loop algorithms require strictly aligned data batches for accurate gradient estimation, the parallel architecture must enforce rigid synchronization. This necessity for coordination across sampling workers inherently compromises the efficiency of data collection, thereby creating a persistent challenge for high-throughput training \cite{qiao2024asmafl, wijmans2019dd, moritz2018ray}.

\begin{figure}[!t]
    \centering
    \includegraphics[width=1\linewidth]{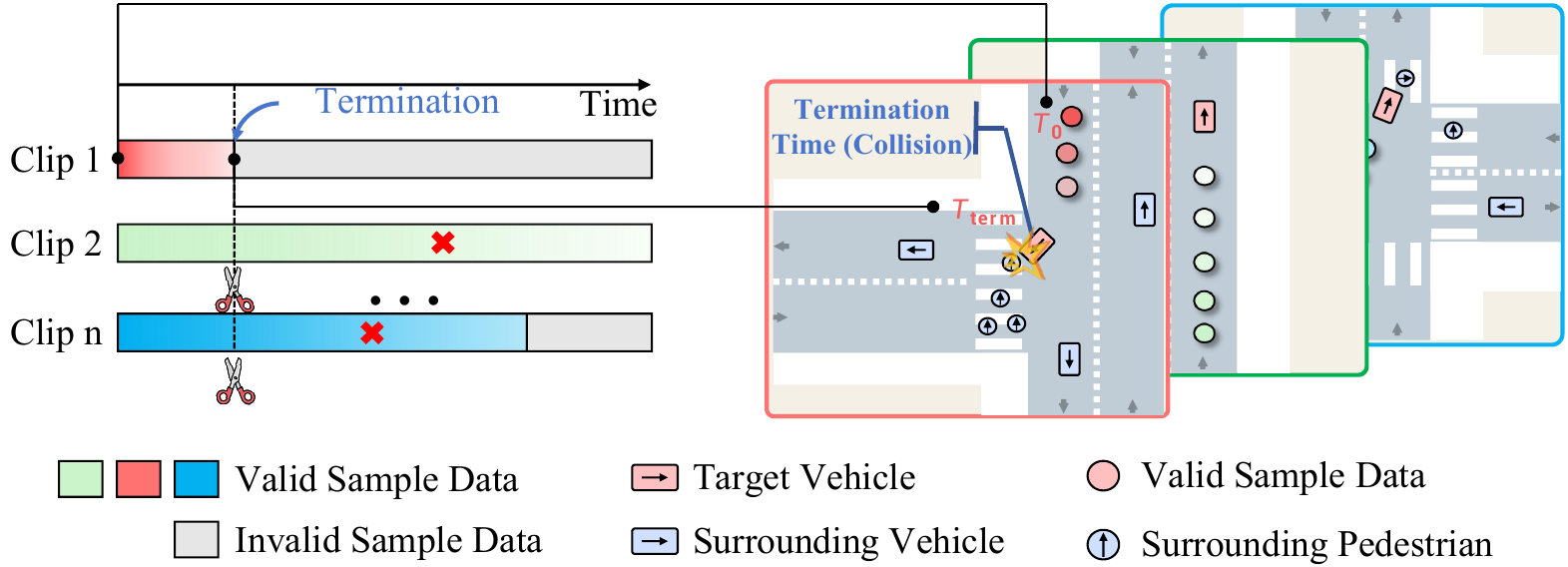}
    \caption{Illustration of the inefficiency caused by the ``reset-on-any-termination'' protocol. The global reset is triggered by the earliest terminating clip (clip 1), forcing the premature truncation of surviving clips. This mechanism discards valuable long-horizon data and increases the frequency of computationally expensive re-initializations.}
    \label{fig:1}
\end{figure}

Nevertheless, the inherent nature of autonomous driving introduces unique complexities, as episode durations are highly stochastic and variable. As illustrated in Figure \ref{fig:1}, while certain clips such as Clip 2, continue to collect valid sample data, a clip such as Clip 1 may terminate prematurely due to a collision. This temporal heterogeneity often necessitates a rigid reset-on-any-termination protocol, where the earliest termination triggers a global reset of all parallel clips. Such resets artificially truncate the termination distribution of the environment, compelling the critic network to systematically underestimate the long-term value of states \cite{pardo2018time,brockman2016openai}. These observations highlight two fundamental challenges that this research seeks to address. At the system level, frequent re-initialization of high-fidelity environments necessitates transferring massive data payloads, including HD maps and sensory states, between hardware hierarchies. This recurrent process creates a severe I/O bottleneck that significantly extends the sampling time, thereby limiting the overall training throughput \cite{kazemkhani2024gpudrive,scheel2022urban}. Simultaneously, at the algorithmic level, the artificial truncation of the termination distribution induces a truncation bias in value estimation, thereby compromising the agent's capacity for credit assignment over extended temporal horizons \cite{pardo2018time,dauner2023parting}.

To accelerate the sampling phase at the system level, the community has optimized simulation engines\cite{vinitsky2022nocturne,zhou2020smarts}. High-performance frameworks, such as EnvPool \cite{weng2022envpool}, Brax \cite{freeman2021brax}, and Isaac Lab \cite{mittal2025isaac}, leverage C++ threading or GPU acceleration to minimize the latency of individual steps. However, while these engines reduce the computation time per step, they do not eliminate the structural overhead caused by rigid synchronization. They typically fail to address the logical alignment delays inherent in episodes of variable length, which means the effective sampling throughput remains capped by the stragglers within the batch, regardless of the raw speed of the engine.

Beyond the optimization of simulation engines, addressing the synchronization bottleneck of variable-length episodes necessitates fundamental interventions at the algorithmic level. However, existing paradigms largely fail to resolve the aforementioned twin challenges: the structural computational idleness caused by rigid resets, and the underestimation of state values that impairs long-horizon forecasting. For instance, while PPG \cite{cobbe2021phasic} enhances sample reuse through phasic decoupling, this approach prioritizes learning efficiency rather than resolving the structural idleness stemming from synchronous waiting during the sampling process. VER \cite{wijmans2022ver} mitigates the straggler effect by collecting non-uniform experience tailored to environment simulation speeds. Despite matching asynchronous throughput in an on-policy manner, the framework remains constrained by individual environment resets. In high-concurrency or frequent-termination scenarios, these operations accumulate latency and restrict overall sampling efficiency. Furthermore, although Seer \cite{qin2025seer} successfully alleviates synchronous long-tail latency, its architecture is specifically tailored to token generation patterns and lacks direct transferability to continuous-state tasks within the domain of autonomous driving. Consequently, the field of autonomous driving currently lacks a sampling mechanism capable of overcoming the simulation bottleneck while maintaining on-policy unbiasedness at the mathematical level \cite{schulman2017proximal,espeholt2019seed}.

This paper addresses these challenges by proposing a novel framework, FAST, which utilizes Dynamic Parallel Sampling Alignment (DPSA) to accelerate parallel sampling using a termination-based strategy. Our approach integrates a virtual continuation mechanism that extends terminated episodes with ``dummy'' data to perform the Scaled Mask-Padding Optimization (SMPO), which effectively bypasses the need for frequent global resets. We demonstrate that decoupling the environment reset from the sampling loop can significantly improve wall-clock sampling efficiency, offering a scalable and robust solution to accelerate large-scale reinforcement learning for autonomous driving applications.

In summary, our main contributions are as follows:

\begin{itemize}
\item At the algorithmic framework level, we propose the DPSA strategy, which leverages a novel space-for-time substitution paradigm to restructure the synchronization process in reinforcement learning. To resolve the severe synchronization bottlenecks and computational idleness inherent in variable-length episodes, DPSA replaces rigid environment resets with a virtual continuation mechanism governed by a delayed termination protocol. Through the virtual continuation of prematurely reset environments, this algorithmic redesign effectively aligns highly heterogeneous dependencies into a structurally uniform temporal horizon, thereby eliminating the structural bias and variance injected by such early terminations while maintaining the theoretical integrity of the sampling distribution.

\item At the policy optimization level, we derive SMPO to mathematically preserve on-policy unbiasedness within the synchronous sampling framework. To address the inherent challenge where auxiliary padding transitions severely distort estimations of the advantage function, SMPO integrates an adaptive loss normalization mechanism. By normalizing the masked loss against the actual number of valid state transitions, this mathematical formulation neutralizes the gradient bias injected by virtual continuations. Consequently, the mechanism enables stable, variance-reduced updates of the policy while completely preserving the strict distribution alignment required for on-policy learning.

\item We conduct an extensive evaluation of the proposed FAST framework using a large-scale, high-fidelity autonomous driving dataset characterized by extreme temporal heterogeneity and complex closed-loop interactions. Experimental results demonstrate that FAST achieves at least a $1.78\times$ speedup in wall-clock time compared to single-clip (SC) baseline, alongside substantially higher effective sampling throughput. Furthermore, by effectively overcoming the straggler effect, our framework avoids premature environment resets, enabling the agent to consistently capture long-horizon interaction data and achieve episode returns comparable to the SC baseline. This validates its efficacy for the development of safety-critical agents in real-world deployment scenarios.

\end{itemize}

The remainder of this paper is organized as follows. Section II provides the formal problem definition in parallel training. Section III details our proposed methodology. Section IV presents the experimental setup, implementation details, and a comprehensive analysis of the results. Finally, Section V concludes the paper.

\section{Problem Definition}

We formulate the autonomous driving task as a Markov Decision Process (MDP), defined by the tuple $(\mathcal{S}, \mathcal{A}, \mathcal{P}, \mathcal{R}, \gamma)$. At each discrete time step $t$, the agent observes a state $s_t \in \mathcal{S}$, executes an action $a_t \in \mathcal{A}$ according to policy $\pi_\theta$, and receives a reward $r_t$. The objective is to maximize the expected cumulative return:

\begin{equation}
    J(\pi) = \mathbb{E}_{\pi,\mathcal{P}} \left[ \sum_{t=0}^{T} \gamma^t r_t \right].
\end{equation}

Crucially, in closed-loop autonomous driving, the episode horizon $T$ is inherently stochastic and highly variable. An episode may terminate prematurely due to safety violations, such as collisions (resulting in a small $T$), or extend until the vehicle reaches a destination (resulting in a large $T$). This temporal heterogeneity presents a fundamental challenge when scaling up training via parallelization.

To address the high sample complexity required for algorithmic convergence, we employ a set of parallel simulation clips, defined as $\mathcal{C} = \{C_1, \dots, C_n\}$.  In many practical synchronous implementations, rollout data are collected in fixed-shape batches for efficient tensorized training. Let $\mathcal{T}_t$ denote the joint transition batch collected at global step $t$:

\begin{equation}
    \mathcal{T}_t = \left\{ \left(s_t^{(i)}, a_t^{(i)}, r_t^{(i)}, s_{t+1}^{(i)}\right) \mid i \in \{1, \dots, n\} \right\}.
\end{equation}

This requirement imposes a rigid synchronization barrier: the global system clock cannot advance to $t+1$ until all $n$ environments have completed their respective transitions and returned the updated state batch.

To formalize this inefficiency and establish the optimization objective, we define the overall effective throughput, $\Phi_{\text{eff}}$, as the number of valid on-policy transitions collected per unit of wall-clock time. The effective throughput can be formulated as the following joint optimization problem:

\begin{equation}
\max \Phi_{\text{eff}} = \Phi_{\text{raw}} \cdot \mu \cdot \eta,
\end{equation}
where the raw sampling throughput $\Phi_{\text{raw}} = N_{\text{total}}/T_{\text{sample}}$ denotes the total number of environment transitions, including both valid and padded samples, generated by the system per unit of wall-clock time; $T_{\text{sample}}$ denotes the wall-clock time required to complete a sampling phase, and $N_{\text{total}}$ denotes the total number of generated samples including padding. The Sample Validity Rate $\mu \in (0, 1]$ represents the proportion of legitimate, on-policy transitions within the total collected samples, defined as $\mu = N_{\text{valid}} / N_{\text{total}}$ with $N_{\text{valid}}$ being the count of valid transitions. Furthermore, the Time Efficiency Ratio $\eta \in (0, 1)$ denotes the proportion of wall-clock time dedicated to actual physical simulation rather than synchronization or reset overhead, expressed as $\eta = T_{\text{step}} / T_{\text{cycle}} = T_{\text{step}} / (T_{\text{step}} + T_{\text{reset}})$, where $T_{\text{step}}$ is the duration of active environment stepping and $T_{\text{cycle}}$ represents the total cycle time including the hardware and physics engine reset latency $T_{\text{reset}}$. This formulation mathematically exposes the core dilemma of existing synchronous engines: maximizing $\Phi_{\text{eff}}$ requires jointly optimizing $\Phi_{\text{raw}}$, $\mu$, and $\eta$.

Specifically, to maintain the tensor shape of $\mathcal{T}_t$ without complex handling, standard strategies often mandate a ``reset-on-any-termination'' protocol, hereafter referred to as Synchronous Global Reset (SGR) \cite{brockman2016openai}. The effective cycle time $T_{\text{cycle}}$ for a batch rollout is strictly governed by the minimum lifespan among the parallel simulation clips:
\begin{equation}
    T_{\text{cycle}} = T_\text{step} + T_\text{reset} = \min \{T_{1}, T_{2}, \dots, T_{n}\} + T_{\text{reset}},
\end{equation}
where $T_{i}$ denotes the duration of the $i$-th environment and $T_{\text{reset}}$ represents the heavy computational overhead of re-initializing the physics engine. This strategy guarantees perfect sample validity ($\mu = 1$). However, it reveals a critical ``frequency bottleneck''. 
As the number of parallel environments $n$ increases, the computational efficiency ratio $\eta$ degrades significantly:
\begin{equation}
    \eta_{\text{SGR}} = \mathbb{E}[\frac{\min_{1 \le i \le n} T_{i}}{\min_{1 \le i \le n}T_{i} + T_{\text{reset}}}].
\end{equation}
Because the probability of at least one environment triggering a premature termination approaches $1$ as $n \to \infty$, the expected minimum lifespan $\mathbb{E}[\min_{1 \le i \le N} T_{i}]$ shrinks drastically. Consequently, $T_{\text{reset}}$ heavily dominates the cycle, substantially reducing $\eta_{\text{SGR}}$ and catastrophically capping the overall throughput $\Phi_{\text{eff}}$. Moreover, as $n$ increases, the growing frequency of resets causes the cumulative reset overhead to increasingly dominate $T_{\text{sample}}$, thereby inflating $T_{\text{sample}}$ and, in turn, degrading $\Phi_{\text{raw}} = N_{\text{total}}/T_{\text{sample}}$ as well.

Conversely, alternative strategies that wait for all environments to finish, such as employing dummy padding, improve $\eta$ but severely reduce the sample validity rate $\mu$, generating wasted computational cycles on useless data.

This drastic reduction in throughput highlights the necessity for a mechanism that decouples the global reset trigger from individual environment terminations. Our proposed method reduces both the frequency and the duration of environment resets, thereby jointly improving $\Phi_{\text{raw}}$ and $\eta$, while keeping the resulting reduction in $\mu$ within a small, controlled margin, to achieve a much higher global $\Phi_{\text{eff}}$.

\begin{figure*}[!t]
    \centering
    \includegraphics[width=1\linewidth]{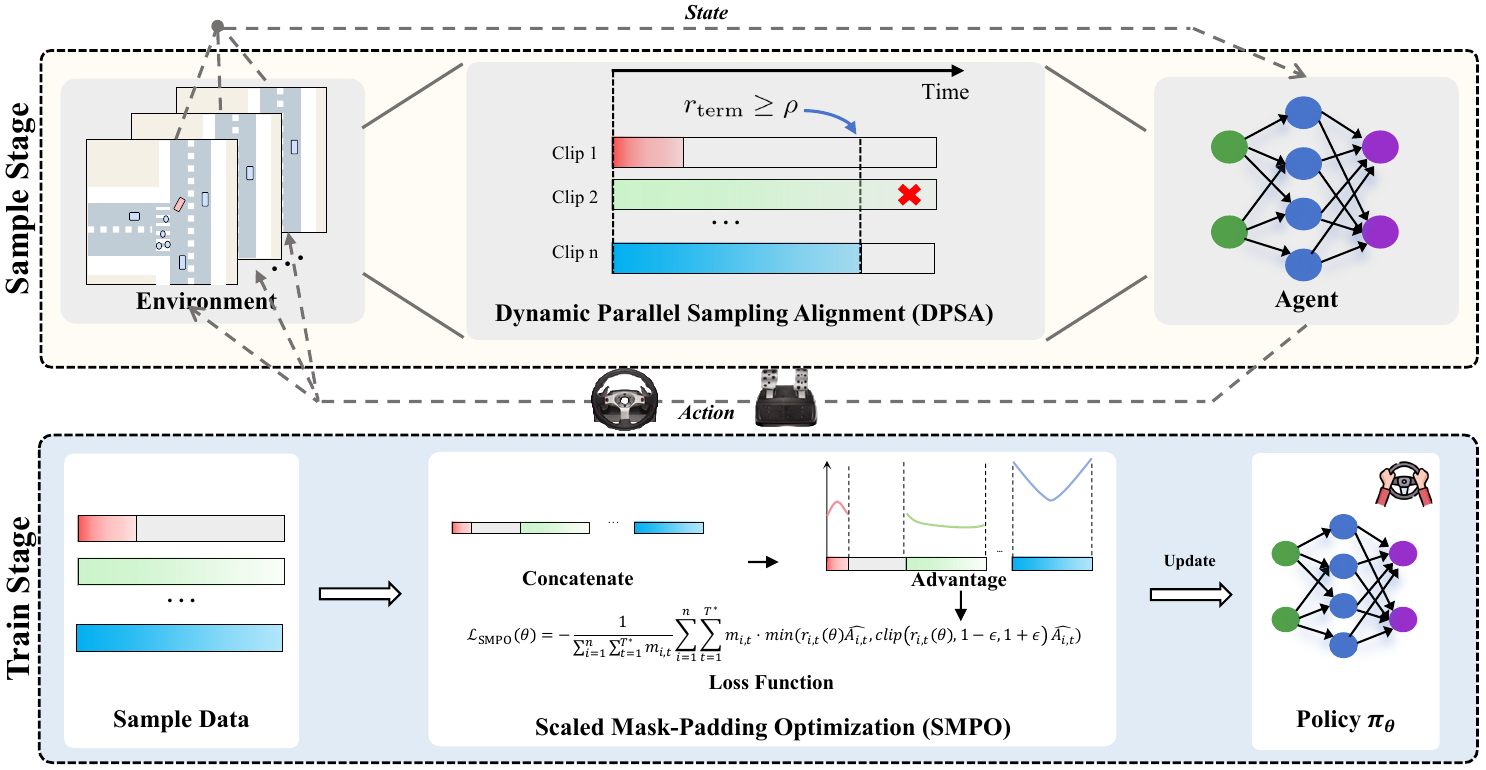}
    \caption{Overview of the FAST framework architecture. The sampling stage implements DPSA, wherein multiple environment clips are executed in parallel and synchronized according to the termination threshold $\tau$. Subsequently, the training stage incorporates the SMPO pipeline; in this phase, rollout data is concatenated and normalized to facilitate the update of the policy $\pi_\theta$ utilizing computed advantages and the objective loss function.}
    \label{fig:2}
\end{figure*}

\section{Methodology}

\subsection{Overview}

Figure \ref{fig:2} illustrates the overall pipeline of the proposed framework, which consists of two primary and interdependent components: the Sample Stage and the Train Stage. During the Sample Stage, a centralized agent interacts with an array of parallel clips to generate state and action data. Within this process, the DPSA mechanism allows terminated clips to enter a virtual continuation state, which satisfies the strict synchronization requirements of the vectorization pipeline. A central monitor dynamically aggregates the completion status and triggers a global truncation once the termination rate of the clips reaches a specific threshold. Subsequently, during the Train Stage, the system concatenates the collected sample data and computes the advantage values. Finally, the framework updates the parameters of the policy through SMPO mechanism, which ensures that the optimization of the policy remains unbiased despite the presence of padded trajectories. Algorithm~\ref{alg:drivesync} summarizes the corresponding execution flow.

\subsection{Dynamic Parallel Sampling Alignment}

To formalize this execution flow, let $\mathcal{C} = \{C_1, \dots, C_n\}$ denote a set of $n$ parallel simulation clips. We define $d_i \in \{0, 1\}$ as a binary completion indicator for clip $C_i$, where $d_i = 1$ signifies that the clip has reached the terminal state. 

\begin{algorithm}[!t]
\small
\caption{Algorithmic Implementation of FAST}
\label{alg:drivesync}
\begin{algorithmic}[1]
\Require Parallel Simulation Clips $\mathcal{C} = \{C_1, \dots, C_n\}$; Initial parameters of the policy $\theta$; Termination threshold $\tau$; Maximum temporal horizon $T_{\mathrm{max}}$.
\Ensure Optimized parameters of the policy $\theta^*$.
\State Initialize $\theta$; \quad $\mathbf{o} \leftarrow \mathrm{Reset}(\mathcal{C})$ 
\Loop
    \Statex \textit{// Sample Stage: Dynamic Parallel Sampling Alignment}
    \State $\mathbf{d} \leftarrow [0]^n$; \quad $\mathbf{T}_{\mathrm{term}} \leftarrow [T_{\mathrm{max}}]^n$; \quad $T^* \leftarrow T_{\mathrm{max}}$
    
    \For{$t = 1$ \textbf{to} $T_{\mathrm{max}}$}
        \State $\mathbf{a} \leftarrow \pi_\theta(\mathbf{o})$
        \State $\mathbf{o}', \mathbf{r}, \mathbf{done} \leftarrow \mathrm{Step}(\mathcal{C}, \mathbf{a})$ \Comment{Execute actions}
        
        \For{$i = 1$ \textbf{to} $n$}
            \If{$\mathrm{done}_{i} = 1$ \textbf{and} $d_i = 0$}
                \State $d_i \leftarrow 1$; \quad $T_{\mathrm{term}}^{(i)} \leftarrow t$
            \EndIf
        \EndFor
        
        \State $\mathbf{o} \leftarrow \mathbf{o}'$
        
        \State $r_{\mathrm{term}} \leftarrow \frac{1}{n} \sum_{i=1}^n d_i$
        \If{$r_{\mathrm{term}} \ge \tau$}
            \State $T^* \leftarrow t$; \quad \textbf{break}
        \EndIf
    \EndFor
    
    \Statex \textit{// Train Stage: Scaled Mask-Padding Optimization}
    \State Compute the temporal validity mask: $m_{i,t} \leftarrow \mathbb{I}(t \le T_{\mathrm{term}}^{(i)})$
    \State Compute advantages $\hat{A}_{i,t}$ via GAE 
    \State Compute gradients $\nabla_\theta \mathcal{L}_\mathrm{SMPO}{}$ using Eq. (9)
    \State Update $\theta$
    
\EndLoop
\end{algorithmic}
\end{algorithm}

The execution workflow commences in the Sample Stage with the DPSA phase, where the agent interacts with parallel simulation clips $\mathcal{C}$ to generate transitions. Unlike standard methods that reset immediately upon individual termination, which thereby causes severe computational bottlenecks, the proposed framework maintains the completion state $d_i$ across the entire batch. Clips that finish early mark $d_i=1$ and record the termination timestamp $T_{\mathrm{term}}^{(i)}$, yet they continue to step synchronously to preserve the integrity of the vectorization pipeline.

Simultaneously, the system dynamically monitors the progress of the clips to optimally trigger a global truncation. The global termination rate, $r_{\mathrm{term}}$, is computed at each step as the proportion of completed clips:
\begin{equation}
    r_{\mathrm{term}}(t) = \frac{1}{n} \sum_{i=1}^n d_i.
\end{equation}
The DPSA phase persists until a global termination signal is triggered at time step $T^*$. As illustrated in the sample stage of Figure \ref{fig:2}, this cutoff is governed by a delayed termination protocol, which acts as a dual-criterion trigger defined by the earliest occurrence of either the satisfaction of the sufficient termination criterion with a threshold $\tau$ (where $r_{\mathrm{term}} \ge \tau$) or the attainment of the pre-defined maximum temporal horizon $T_{\mathrm{max}}$:
\begin{equation}
    T^* = \min \{ t \mid r_{\mathrm{term}}(t) \ge \tau \lor t = T_{\mathrm{max}} \}.
\end{equation}
Upon reaching $T^*$, all remaining active clips are truncated to release the synchronization barrier, thereby maintaining high computational throughput by pruning the unnecessary tail latency caused by exceptionally long episodes.

To preserve the batch alignment for clips that terminate prematurely, we introduce a temporal validity mask $m_{i,t}$ to ensure that the resulting pseudo-transitions do not bias the gradient of the policy:
\begin{equation}
    m_{i,t} = \begin{cases} 
    1, & \text{if } t \le T_{\mathrm{term}}^{(i)} \\ 
    0, & \text{if } t > T_{\mathrm{term}}^{(i)}. 
    \end{cases}
\end{equation}

\begin{figure}[!t] 
    \centering     
    \includegraphics[width=0.9\linewidth]{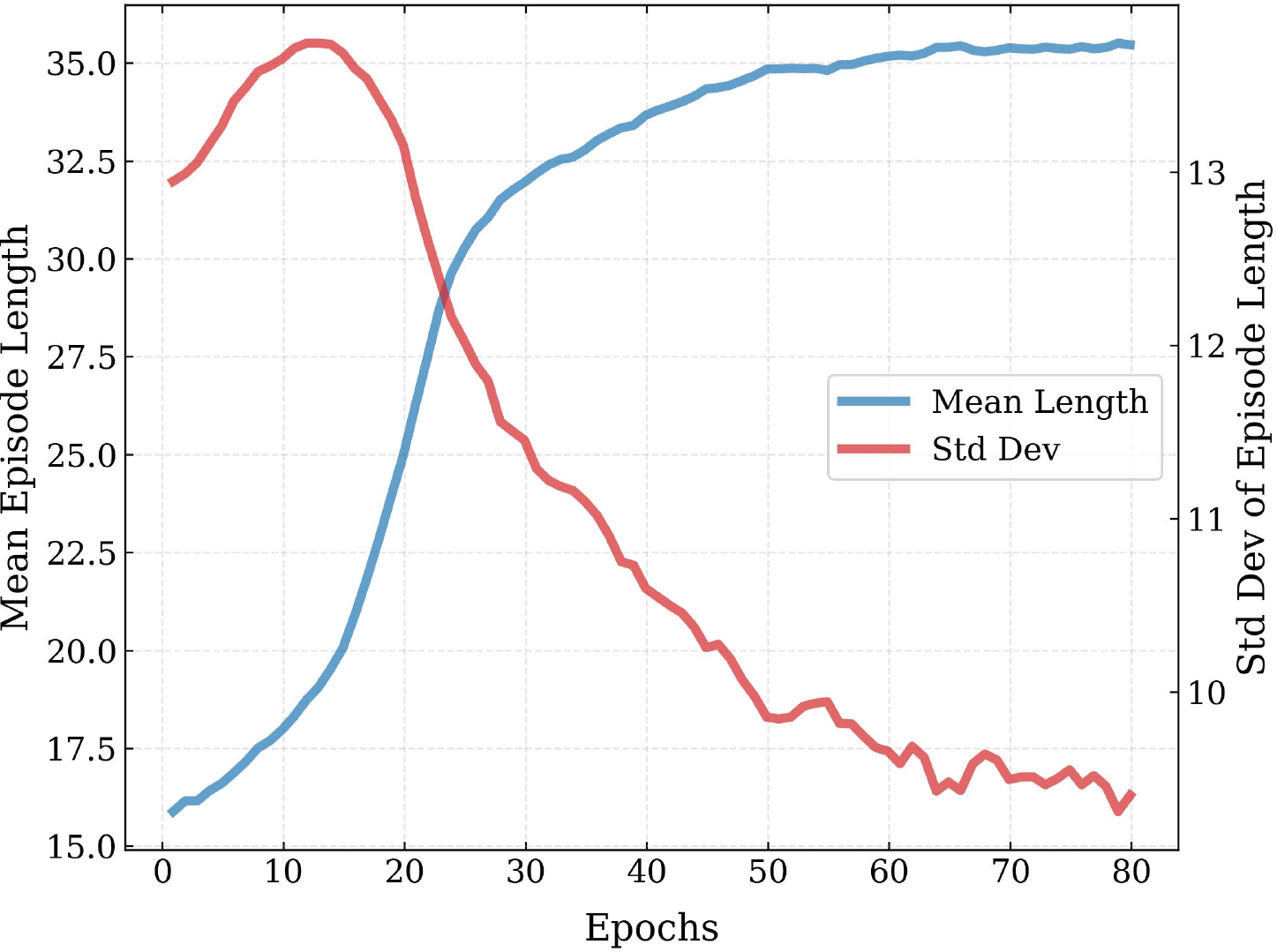}
    \caption{Statistics of the episode lengths derived from SC baseline.}

    \label{fig:3}
\end{figure}

\subsection{Scaled Mask-Padding Optimization}

As illustrated in the Train Stage of the workflow, following the global concatenation of the collected sample data, the system computes the generalized advantage estimation within each independent clip. To precisely align with the stepping of the underlying vectorized environment, the advantage function $\hat{A}_{i,t}$ for any time step $t$ in clip $C_i$ is calculated via backward recursion:\begin{equation}\hat{A}_{i,t} = \delta_{i,t} + \gamma^{k_{i,t}} \lambda (1 - d_{i,t}) \hat{A}_{i,t+1}
\end{equation}
where the single-step temporal difference error is defined as $\delta_{i,t} = r_{i,t} + \gamma^{k_{i,t}} (1 - d_{i,t}) V(s_{i,t+1}) - V(s_{i,t})$. Due to the blocking mechanism of the termination indicator $d_{i,t}$, virtual steps executed after $T_{\mathrm{term}}^{(i)}$ do not corrupt the advantage values of preceding valid state transitions. Furthermore, value bootstrapping is naturally achieved at the global truncation point $T^*$.

Upon completing the exact calculation of advantage values, the system computes the loss and performs policy updates based on the concatenated data. An inherent challenge of this continuous alignment method is mitigating the gradient variance induced by fluctuating valid sample densities. To achieve an unbiased empirical estimate of the policy gradient and ensure optimization stability, we formulate the SMPO objective function by normalizing the masked loss with the actual count of valid transitions:

\begin{footnotesize}
\begin{equation}
\begin{aligned}
\mathcal{L}_{\mathrm{SMPO}}(\theta) = 
- \frac{1}{\sum_{i=1}^n \sum_{t=1}^{T^*} m_{i,t} } \sum_{i=1}^n \sum_{t=1}^{T^*} m_{i,t} \cdot \min \Big( &\rho_{i,t}(\theta) \hat{A}_{i,t}, 
\\ \text{clip}(\rho_{i,t}(\theta), 1-\epsilon, 1+\epsilon) \hat{A}_{i,t} \Big),
\end{aligned}
\end{equation}
\end{footnotesize}\noindent where $\theta$ denotes the parameters of the policy, and $\mathbb{E}_{k}$ represents the empirical expectation over the data collected at iteration $k$. The term $\rho_{i,t}(\theta) = \pi_\theta(a_{i,t}|s_{i,t})/\pi_{\theta_{\text{old}}}(a_{i,t}|s_{i,t})$ represents the probability ratio between the new policy and the old policy, $\hat{A}_{i,t}$ is the estimated advantage value, and $\epsilon$ serves as the clipping hyperparameter. Consequently, by scaling the loss with the total volume, the framework ensures that the optimization step size remains invariant to the sparsity of the data, which effectively nullifies the noise from padding while maintaining robust convergence properties that are equivalent to standard training with a fixed horizon. After computing the loss, the system updates the parameters of the policy accordingly to complete the current iteration cycle.

\begin{figure*}[!ht] 
    \centering     
    \includegraphics[width=0.95\linewidth]{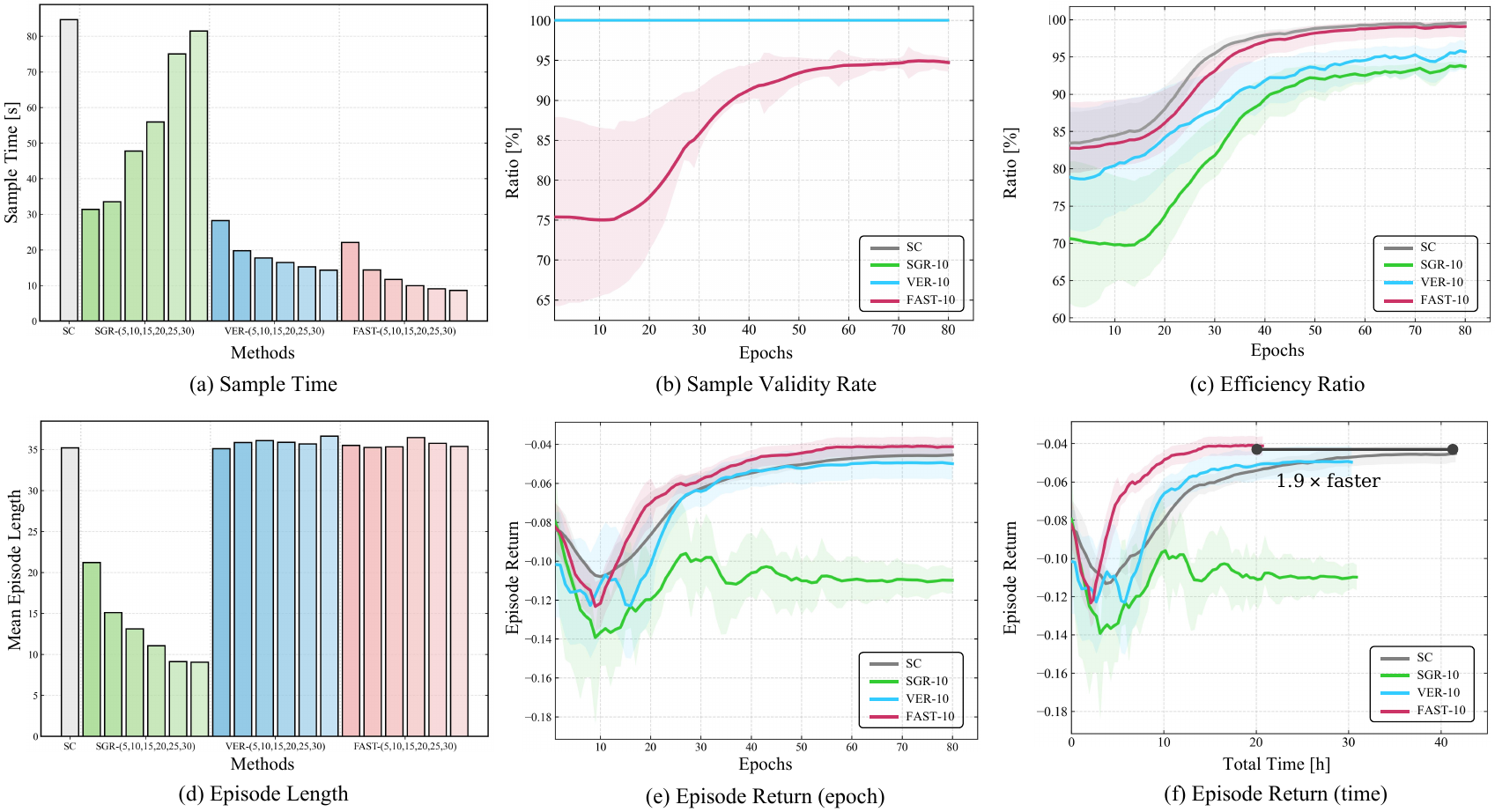}
    \caption{Comparative Evaluation of the SC Baseline, SGR, VER, and FAST Frameworks. The shaded region represents a 95\% confidence interval over 3 independent seeds.}
    \label{fig:5}
\end{figure*}

\begin{figure}[!t] 
    \centering     
    \includegraphics[width=0.95\linewidth]{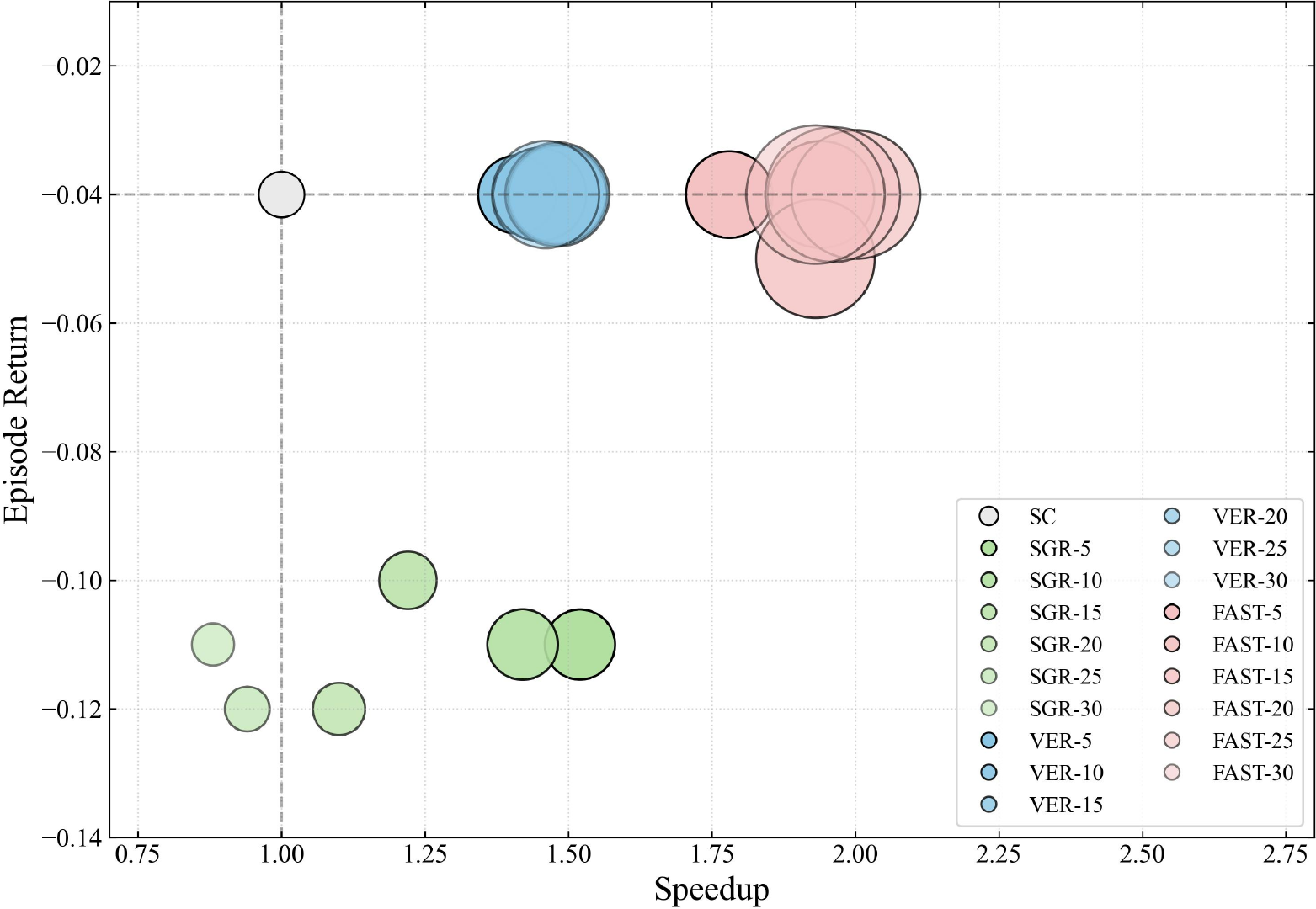}
    \caption{Comprehensive trade-off analysis between execution speedup and algorithmic stability. Bubble dimensions denote the effective throughput ($\Phi_{\text{eff}}$),  illustrating the performace of the SC baseline, SGR, VER and FAST under varying configurations.}

    \label{fig:4}
\end{figure}

\section{Experiments}
\subsection{Dataset}

We validate our framework on a large-scale, proprietary autonomous driving dataset from DiDi, which specifically curates high-difficulty ``takeover'' scenarios involving human intervention. In this setup, the maximum sampling horizon, defined as the total episode length, is strictly capped at 40 steps. Crucially, however, the actual temporal duration of individual data clips exhibits extreme variance, ranging from split-second emergency maneuvers to extended cruising segments. Because of the dataset's inherent difficulty, baseline RL agents experience a significantly high frequency of early terminations, such as collisions or safety envelope violations. As illustrated in Figure \ref{fig:3}, this temporal heterogeneity is empirically evident throughout the training process. During the initial epochs, the agent exhibits a low mean episode length coupled with a high standard deviation, indicating significant variance and frequent early terminations. Although the policy gradually stabilizes as training progresses, which is evidenced by the mean episode length increasing to a plateau near the theoretical maximum alongside a consistent decrease in the standard deviation, a notable disparity in clip durations persists. If evaluated using a traditional synchronous engine, this highly dispersed distribution of episode lengths perfectly exemplifies the straggler bottleneck: rapid terminations in challenging, short clips would continuously trigger the global reset protocol, causing the time efficiency ratio $\eta$ to catastrophically collapse. Therefore, this dataset serves as an ideal, rigorous testbed to demonstrate our method's ability to maintain high throughput despite severe temporal heterogeneity.

\subsection{Experiment Setup}

The training process is executed on a high-performance computing node equipped with 8 NVIDIA H20 GPUs. To standardize the training loop against the extreme variance of episode lengths, we define a training epoch strictly by sample volume rather than the number of completed driving clips. Specifically, each training epoch entails collecting a rollout buffer of 1,000 environment transitions. Based on this constant volume, the raw sampling throughput of the system is formally defined as $\Phi_{\text{raw}} = 1000 / T_{\text{sample}}$, where $T_{\text{sample}}$ represents the time required to complete the sampling phase. Once this data collection phase is complete, we perform 5 consecutive rounds of optimization updates over the buffer, utilizing a global batch size of 2560 distributed across the devices.  To mitigate gradient estimation variance during these updates, Generalized Advantage Estimation is applied with a discount factor $\gamma=0.8$ and a smoothing parameter $\lambda=0.7$.

\begin{table*}[t]
\centering
\caption{Quantitative evaluation of computational efficiency and policy performance across varying parallel scales. The time metrics $T_{\text{sample}}$, $T_{\text{train}}$, and $T_{\text{total}}$ denote the mean values per training epoch, while the throughput $\Phi_{\text{raw}}$ and $\Phi_{\text{eff}}$, sample validity rate $\mu$, time efficiency ratio $\eta$, and Episode Return represent the final values recorded at the conclusion of the training process. Bold values indicate the best performance within each parallel scale group. \textcolor{blue!50}{Purple} highlights the key metrics.}
\label{table:1}

\setlength{\aboverulesep}{0pt}
\setlength{\belowrulesep}{0pt}
\setlength{\tabcolsep}{3.3mm} 
\renewcommand{\arraystretch}{1.4} 

\colorlet{mychoice}{blue!8} 

\resizebox{0.95\textwidth}{!}{
\begin{tabular}{
    c | l | 
    c 
    c 
    c 
    >{\columncolor{mychoice}}c |
    c 
    c     
    c
    >{\columncolor{mychoice}}c |    
    >{\columncolor{mychoice}}c    
}
\toprule
Clips & Method & $T_{\text{sample}}$ (s) & $T_{\text{train}}$ (s) & $T_{\text{total}}$ (s) & Speedup ($\times$) & $\mu$ (\%) & $\eta$ (\%) & $\Phi_{\text{raw}}$ & $\Phi_{\text{eff}}$ & Episode Return \\ 
\midrule
1 & SC \cite{schulman2017proximal} & 84.67 & 68.92 & 189.68 & 1.00 & 100.00 & 99.95 & 11.81 & 11.80 & -0.04 \\ 
\midrule

\multirow{3}{*}{5} 
& SGR-5 \cite{brockman2016openai} & 31.34 & 69.39 & 124.86 & 1.52 & \textbf{100.00} & 95.64 & 31.91 & 30.52 & -0.11 \\ 
& VER-5 \cite{wijmans2022ver}  & 28.25 & 69.84 & 134.70 & 1.41 & \textbf{100.00} & 97.62 & 35.40 & 34.56 & \textbf{-0.04} \\ 
& \textbf{FAST-5} & \textbf{22.07} & \textbf{68.27} & \textbf{106.47} & \textbf{1.78} & 93.63 & \textbf{99.90} & \textbf{45.31} & \textbf{42.38} & \textbf{-0.04} \\ 
\midrule

\multirow{3}{*}{10} 
& SGR-10 \cite{brockman2016openai} & 33.51 & 68.72 & 133.77 & 1.42 & \textbf{100.00} & 93.03 & 29.84 & 27.76 & -0.11 \\ 
& VER-10 \cite{wijmans2022ver} & 19.73 & 68.67 & 130.81 & 1.45 & \textbf{100.00} & 96.45 & 50.68 & 48.88 & -0.05 \\ 
& \textbf{FAST-10} & \textbf{14.35} & \textbf{67.79} & \textbf{97.61} & \textbf{1.94} & 94.61 & \textbf{99.83} & \textbf{69.69} & \textbf{65.82} & \textbf{-0.04} \\ 
\midrule

\multirow{3}{*}{15} 
& SGR-15 \cite{brockman2016openai}  & 47.71 & \textbf{68.38} & 155.92 & 1.22 & \textbf{100.00} & 89.33 & 20.96 & 18.72 & -0.10 \\ 
& VER-15 \cite{wijmans2022ver} & 17.72 & 68.82 & 127.84 & 1.48 & \textbf{100.00} & 94.42 & 56.43 & 53.28 & \textbf{-0.04} \\ 
& \textbf{FAST-15} & \textbf{11.67} & 68.16 & \textbf{98.17} & \textbf{1.93} & 93.02 & \textbf{99.83} & \textbf{85.69} & \textbf{79.57} & -0.05 \\ 
\midrule

\multirow{3}{*}{20} 
& SGR-20 \cite{brockman2016openai} & 55.95 & 69.19 & 171.99 & 1.10 & \textbf{100.00} & 86.52 & 17.87 & 15.46 & -0.12 \\ 
& VER-20 \cite{wijmans2022ver} & 16.47 & \textbf{68.53} & 128.46 & 1.48 & \textbf{100.00} & 93.68 & 60.72 & 56.88 & \textbf{-0.04} \\ 
& \textbf{FAST-20} & \textbf{9.95} & 68.61 & \textbf{94.85} & \textbf{2.00} & 93.46 & \textbf{99.72} & \textbf{100.50} & \textbf{93.67} & \textbf{-0.04} \\ 
\midrule

\multirow{3}{*}{25} 
& SGR-25 \cite{brockman2016openai} & 75.04 & \textbf{68.62} & 201.76 & 0.94 & \textbf{100.00} & 84.66 & 13.33 & 11.28 & -0.12 \\ 
& VER-25 \cite{wijmans2022ver} & 15.23 & 68.99 & 128.37 & 1.48 & \textbf{100.00} & 93.44 & 65.66 & 61.35 & \textbf{-0.04} \\ 
& \textbf{FAST-25} & \textbf{9.05} & 68.63 & \textbf{96.90} & \textbf{1.96} & 93.10 & \textbf{99.72} & \textbf{110.50} & \textbf{102.98} & \textbf{-0.04} \\ 
\midrule

\multirow{3}{*}{30} 
& SGR-30 \cite{brockman2016openai} & 81.50 & \textbf{68.69} & 215.07 & 0.88 & \textbf{100.00} & 81.97 & 12.27 & 10.06 & -0.11 \\ 
& VER-30 \cite{wijmans2022ver} & 14.27 & 69.32 & 129.48 & 1.46 & \textbf{100.00} & 92.69 & 70.08 & 64.95 & \textbf{-0.04} \\ 
& \textbf{FAST-30} & \textbf{8.59} & 69.10 & \textbf{98.26} & \textbf{1.93} & 92.35 & \textbf{99.67} & \textbf{116.41} & \textbf{107.15} & \textbf{-0.04} \\ 
\bottomrule
\end{tabular}
}
\end{table*}

The policy network is instantiated as a Transformer-based planner. It undergoes fine-tuning for 80 epochs using the AdamW optimizer (learning rate=$1 \times 10^{-5}$, weight decay=0.01), where a ConstantLR scheduler is adopted to ensure the stability of the pre-trained backbone. Regarding the simulation environment, we configure a 3-step rollout horizon per iteration, incorporating the nearest 64 background agents to capture complex interactions. Additionally, the clipping parameter is set to $\epsilon=0.1$ with a critic loss coefficient of 1.0.

Finally, the reward function is meticulously formulated to balance multiple objectives: \textit{Safety} by penalizing collisions and hard braking, \textit{Compliance} with traffic lights and speed limits, \textit{Navigation} for route adherence, and \textit{Comfort} via jerk minimization. These distinctive components are aggregated as a weighted linear combination to constitute the total reward.

\subsection{Evaluation Result}

We evaluate the performance of the proposed FAST framework alongside several highly competitive baselines, including SGR \cite{brockman2016openai} and VER \cite{wijmans2022ver}, across varying scales of parallel environments. This evaluation is conducted from three distinct perspectives: execution time, effective throughput, and algorithmic stability. The comprehensive results are detailed in Table \ref{table:1} and Figure \ref{fig:5}. Specifically, Figure \ref{fig:4} presents a synthetic evaluation across the three aforementioned perspectives. Overall, the FAST framework consistently achieves optimal performance across all clip configurations, thereby demonstrating the inherent superiority of the proposed method.

\subsubsection{Execution Time} As illustrated in Table \ref{table:1}, all evaluated methods maintain a stable training duration $T_{\text{train}}$ of approximately 68s across various parallel configurations. Furthermore, the FAST framework exhibits a substantial improvement in sampling efficiency compared to the SC benchmark. Specifically, as shown in \ref{fig:4}  (a), the sampling time $T_{\text{sample}}$ for FAST-10 is reduced to 14.35s, representing an 83.0\% decrease compared to the 84.67s of the SC baseline and achieving a 1.94$\times$ overall speedup. This significantly outperforms the speedups of SGR-10 (1.42$\times$) and VER-10 (1.45$\times$). In contrast, the sampling performance of the SGR algorithm undergoes degradation as the number of parallel environments increases. Under the 30-clip configuration, the speedup of SGR further drops to 0.88$\times$, rendering the method slower than the non-parallelized baseline. While VER maintains a relatively steady efficiency gain, the inherent overhead of frequent data acquisition in its asynchronous design prevents it from surpassing the performance of FAST. Notably, $T_{\text{total}}$ slightly exceeds the arithmetic sum of $T_{\text{sample}}$ and $T_{\text{train}}$. This discrepancy stems from serial auxiliary overheads, such as diagnostic visualization and explicit policy weight synchronization, which are executed within the training loop but excluded from the two independent timers, thus manifesting as additional latency in the final $T_{\text{total}}$ profile.
\begin{table*}[!t]
\centering
\caption{Evaluation results of the closed-loop metrics for the FAST-10 framework and the SC baseline. The variables $N^{+}$ and $N^{-}$ denote the counts of positive Fail-to-Pass transitions and negative Pass-to-Fail regressions, respectively. A 95\% confidence interval ($\mathcal{I}_{CI}$) encompassing zero indicates the absence of systematic performance degradation.}
\label{tab:2}
\resizebox{0.85\textwidth}{!}{
\begin{tabular}{ccccccc}
\toprule
 Category& Metric& Meaning& Total Samples& $N^{+}$ & $N^{-}$ &$\mathcal{I}_{CI}$ \\ \midrule
\multirow{3}{*}{Safety} 
 & $S_{\text{coll}}$ & The true degree of collision
 & 4616  & 31 & 38 & [-0.20\%, 0.51\%] \\
 & $S_{\text{risk}}$ & Evaluate leading agents risk & 529 & 4 & 6 & [-0.80\%, 1.56\%] \\
 & $S_{\text{safe}}$ & Violation of safety boundary margins & 300 & 0 & 3 & [-0.13\%, 2.13\%] \\ \midrule
\multirow{2}{*}{Comfort} 
 & $C_{\text{brake}}$ & The likelihood of a harsh brake & 344 & 5 & 8 & [-1.19\%, 2.94\%] \\
 & $C_{\text{acc}}$ & Abrupt or excessive acceleration & 197 & 8 & 8 & [-4.20\%, 4.20\%] \\ \midrule
\multirow{6}{*}{Legality} 
 & $L_{\text{line}}$ & Whether the car is over solid line
 & 367 & 8 & 15 & [-0.66\%, 4.50\%] \\
 & $L_{\text{red}}$ & Failure to stop at red signals & 696 & 1 & 0 & [-0.43\%, 0.14\%] \\
 & $L_{\text{yellow}}$ & Improper entry during yellow signals & 696 & 5 & 1 & [-1.26\%, 0.11\%] \\
 & $L_{\text{park}}$ & Accuracy of parking spot positioning & 696 & 6 & 8 & [-0.77\%, 1.34\%] \\
 & $L_{\text{nudge}}$ & Precision of lateral bypass nudges & 1135 & 2 & 0 & [-0.42\%, 0.07\%] \\
 & $L_{\text{speed}}$ & Deviation from speed regulations & 377 & 2 & 1 & [-1.18\%, 0.64\%] \\ \midrule
\multirow{3}{*}{Efficiency} 
 & $E_{\text{speed}}$ & Ratio of ego speed to max speed limit & 396 &  9 & 4 & [-3.05\%, 0.52\%] \\
 & $E_{\text{stuck}}$ & Total stuck time & 51  & 2 & 2 & [-7.76\%, 7.76\%] \\
 & $E_{\text{dist}}$ & Pull-out distance deviation ratio & 51 & 1 & 1 & [-5.49\%, 5.49\%] \\ \bottomrule
\end{tabular}
}
\end{table*}
\subsubsection{Effective Throughput} 
The computational efficiency of the model is evaluated through the effective throughput $\Phi_{\text{eff}}$, which is fundamentally governed by the joint interplay among $\Phi_{\text{raw}}$, the sample validity rate $\mu$, and the time efficiency ratio $\eta$. As detailed in Table \ref{table:1} and Figure \ref{fig:5} (b) and (c), although both SGR and VER maintain a perfect sample validity rate of $\mu = 100.00\%$, the high frequency of environment resets inherent in these methods leads to a significant degradation in the time efficiency ratio $\eta$. In contrast, while the FAST framework does not sustain a perfect $\mu$ initially, the sample validity rate of the method consistently converges to at least 92\% during the training process. Furthermore, by substantially reducing the frequency and duration of global resets, FAST maintains a near-perfect $\eta$ of over 99.6\% across all parallel configurations while jointly improving $\Phi_{\text{raw}}$. Consequently, the final effective throughput of FAST far exceeds the performance of SGR and VER, reaching a peak $\Phi_{\text{eff}}$ of 107.15 under the 30-clip configuration, which represents a 9.08-fold improvement over the SC benchmark. These results demonstrate that jointly improving $\Phi_{\text{raw}}$ and $\eta$ through reduced reset frequency and duration, rather than trading $\mu$ for $\eta$ alone, is the primary driver for achieving high-throughput data acquisition in large-scale distributed training.
\subsubsection{Episode Length} 
The statistical distributions of episode lengths for the three methods under various configurations are illustrated in Figure~\ref{fig:5} (d). Empirical results demonstrate that the mean episode length of the SGR baseline undergoes a precipitous decline as the scale of parallelism increases. This phenomenon is attributable to the rigid synchronization mechanism, wherein the termination of any individual environment necessitates an immediate global reset. Such a protocol inevitably truncates active episodes within other concurrent environments, thereby inducing a severe short-horizon bias and preventing the agent from acquiring the diverse, long-horizon data essential for effective learning. In contrast, both the VER and FAST frameworks consistently maintain a stable mean duration of approximately 35 steps, aligning with the SC baseline. This stability stems from the decoupling of the global reset trigger from individual terminations, which eliminates short-horizon bias and facilitates more robust policy optimization.

\subsubsection{Algorithmic Stability}
As summarized in Figure \ref{fig:5} (e), the algorithmic stability of FAST is evaluated alongside SGR and VER. Throughout the training process, the episodic returns of both FAST and VER align closely with the SC benchmark of $-$0.04, effectively avoiding the performance degradation observed in the synchronized baseline. In contrast, the performance of the SGR algorithm exhibits a continuous decline, with the episodic return dropping significantly to approximately $-$0.12. This failure is primarily attributed to the high frequency of global resets, which leads to the premature termination of episodes and prevents the agent from capturing critical long-horizon data. Consequently, the resulting short-horizon bias severely impairs the ability of the model to learn complex credit assignments over extended temporal sequences. These results demonstrate that the proposed architecture is particularly effective in addressing the structural limitations of previous models, ensuring that high concurrency does not compromise the stability of the learning process.

\subsubsection{Training efficiency}
As illustrated in Figure \ref{fig:5} (f), FAST-10 achieves a superior convergence rate compared to all evaluated baseline configurations. Specifically, the FAST-10 model reaches a stable performance plateau of approximately $-0.05$ within only about 20 hours of training time. In contrast, the standard SC baseline requires over 40 hours to attain a comparable return level, while the accelerated variants SGR-10 and VER-10 necessitate approximately 30 hours to reach convergence. This significant reduction in temporal overhead corresponds to a $1.9\times$ speedup in training efficiency relative to the SC baseline, as highlighted by the experimental markers. The rapid initial ascent observed in FAST-10 suggests a more effective utilization of computational resources, enabling the agent to bypass the prolonged exploration phases that typically hinder conventional reinforcement learning methods.

\subsection{Closed-loop Simulation Evaluation}
Closed-loop evaluation is indispensable, as cumulative episode return often serves as an imperfect proxy for actual driving performance. A policy that achieves high returns within short, isolated training clips may nevertheless fail during continuous simulations, in which planning errors tend to compound over time. However, a direct comparison of mean metric values is insufficient to distinguish genuine performance shifts from the stochastic noise inherent in complex simulations. To provide a rigorous proof of policy fidelity, an evaluation of statistical equivalence, grounded in the Jackknife resampling principle\cite{sawyer2005resampling,hansen2012jackknife}, is conducted across 14 comprehensive metrics categorized into four pivotal dimensions: safety, driving experience, compliance, and efficiency.

\begin{figure}[!t] 
    \centering     
    \includegraphics[width=0.95\linewidth]{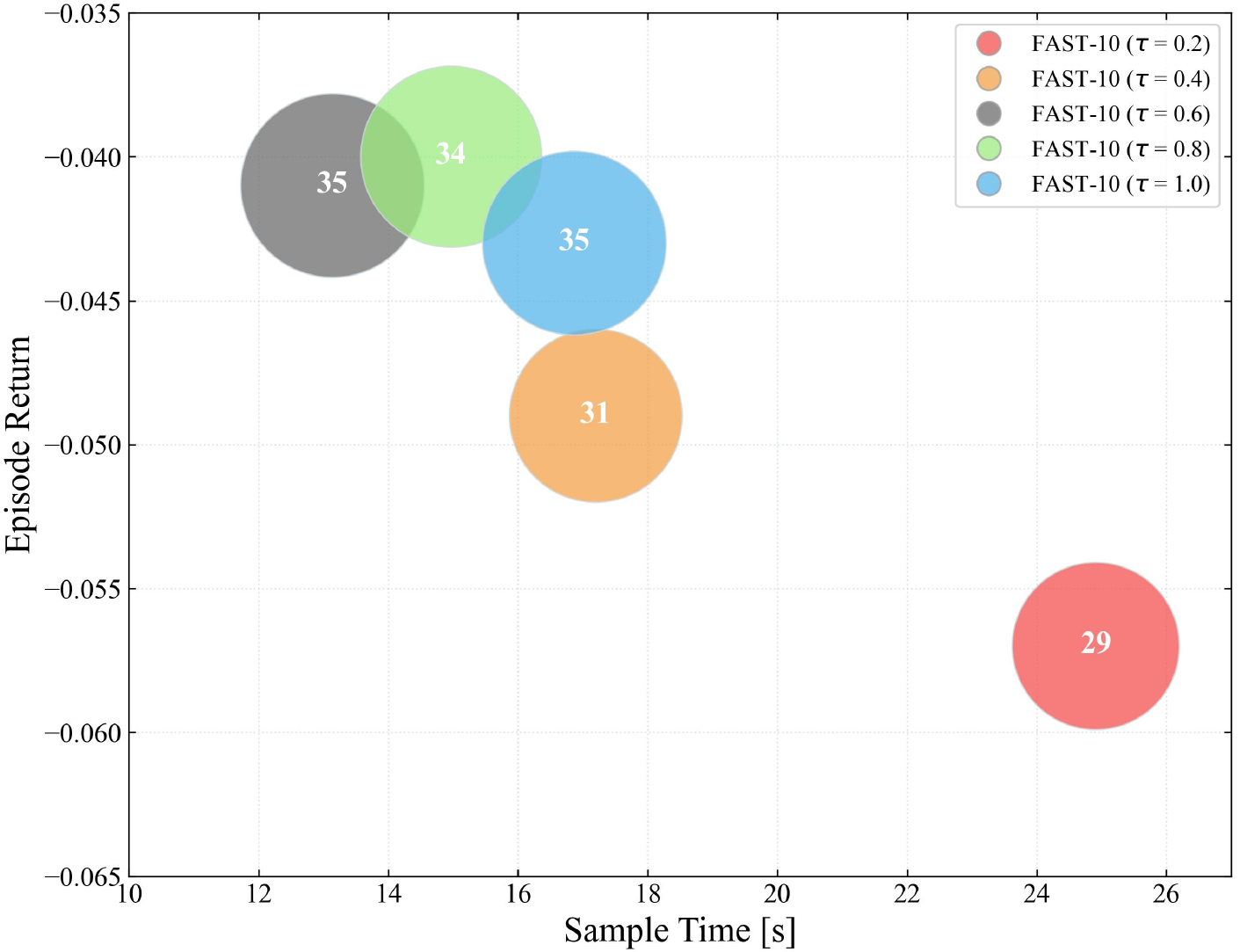}
    \caption{Comprehensive trade-off analysis between sampling time and episode return. The diameters of the bubbles represent the mean episode length, demonstrating the performance of the FAST-10 framework across varying threshold ($\tau$) configurations.}

    \label{fig:6}
\end{figure}

To evaluate the performance of the proposed framework, a discrete scoring formulation is established. For each evaluation metric, a corresponding data subset is utilized, wherein each sample is assigned a discrete value $x \in \{1, 0, -1\}$ to encode the relative transition in metric status. Specifically, with the sequential baseline serving as the reference method, a value of $x = 1$ denotes a positive Fail-to-Pass transition ($N^{+}$), indicating that the parallel agent succeeds while the baseline fails. Conversely, a value of $x = -1$ represents a negative Pass-to-Fail regression ($N^{-}$), and $x = 0$ indicates that the evaluation outcome remains identical. Furthermore, the 95\% confidence interval ($\mathcal{I}_{CI}$) of the mean difference is utilized to assess whether the observed discrepancies exhibit statistical significance. The specific calculation is formulated as follows:

\begin{equation}
\mathcal{I}_{CI} = \mu_X \pm 1.96 \cdot \frac{\sigma_X}{\sqrt{n-1}}
\end{equation}where $\mu_X$ represents the mean of the discrete transition distribution $X$, while $\sigma_X$ denotes the sample standard deviation of these scores. The parameter $n$ signifies the total number of evaluation samples. The core criterion for establishing a conclusion of statistical equivalence rests on whether this 95\% confidence interval encompasses the zero point. An interval straddling zero implies that at a 95\% confidence level, the observed numerical fluctuations, even if discrete transition cases from Fail-to-Pass or from Pass-to-Fail exist, are statistically indistinguishable from random noise rather than systematic degradation. As summarized in Table \ref{tab:2}, the FAST 10 configuration exhibits high consistency with the SC baseline across all evaluated metrics. This statistical evidence confirms that the parallel mechanism achieves a substantial reduction in sampling time without introducing any performance degradation. Ultimately, the profile of statistical equivalence across all categories corroborates that the FAST framework effectively scales training throughput while maintaining the highest level of policy fidelity.

\subsection{Ablation Study on Threshold $\tau$}

\begin{figure}[!t] 
    \centering     
    \includegraphics[width=0.95\linewidth]{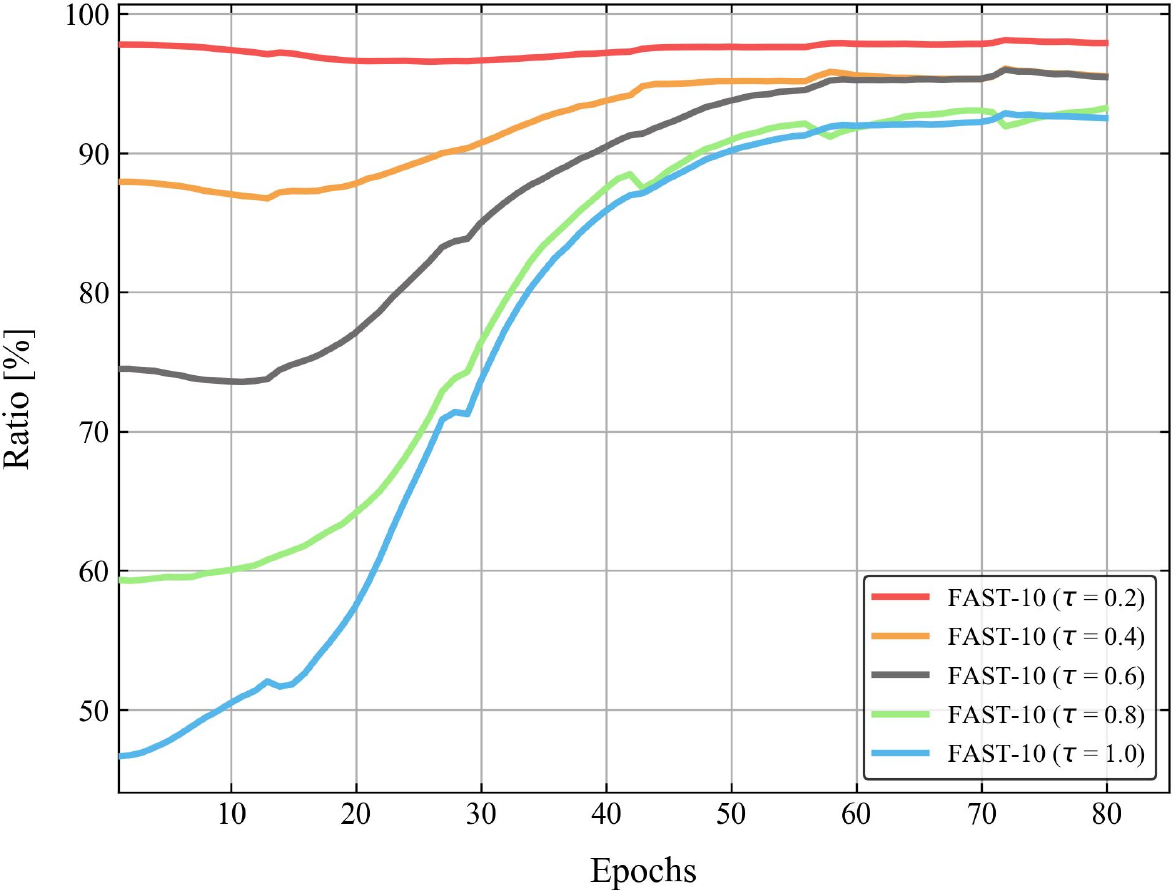}
    \caption{Evolution of Sample Validity Rate $\mu$ over Training Epochs for FAST-10 with Varying $\tau$ Values.}

    \label{fig:7}
\end{figure}

The threshold $\tau$ serves as a critical hyperparameter that modulates the delicate balance between synchronization latency and data validity. To determine the optimal configuration for this parameter, we conducted experiments across various settings, as illustrated in Figure \ref{fig:6}. The ideal parameterization is characterized by a high cumulative episode return, minimized sampling time, and a stable mean episode length that aligns with SC baseline of approximately 35 steps. Empirical results indicate that $\tau = 0.6$ represents the optimal value. As the value of $\tau$ decreases, the framework progressively regresses toward the SGR protocol. This regression is expected, as a threshold of $\tau = 0$ effectively recovers the "reset-on-any-termination" condition of the SGR baseline. Conversely, increasing the value of $\tau$ beyond the optimal point provides no performance benefit; instead, it prolongs the synchronization overhead, thereby increasing the overall sampling time and reducing the sampling speed of the model.
Figure \ref{fig:7} depicts the evolution of the Sample Validity Rate $\mu$ during the training process for different values of $\tau$. While a low threshold such as $\tau = 0.2$ initially yields a high validity rate, this setting causes the system to degenerate into a synchronous global reset pattern. Such a degeneration results in a catastrophic reduction in episode length and impairs learning performance. In contrast, the configuration with $\tau = 0.6$ maintains trajectory integrity while achieving the highest final Sample Validity Rate. Consequently, $\tau = 0.6$ constitutes the optimal balance between sampling efficiency and policy fidelity.

\section{Conclusion}

This paper introduces FAST, a novel synchronous parallel training framework that utilizes DPSA to accelerate large-scale reinforcement learning within high-fidelity simulation environments. The proposed framework significantly enhances the throughput of effective sampling for episodes of variable length. Furthermore, an innovative SMPO mechanism extends terminated episodes through the use of dummy data, thereby obviating the necessity for frequent global resets. This approach effectively addresses the simulation bottlenecks and temporal instability inherent in traditional synchronous methods, facilitating the efficient optimization of autonomous driving policies. Comprehensive evaluations demonstrate that FAST outperforms established baseline methods by achieving substantial speedups and a marked reduction in sampling latency without compromising asymptotic performance. In summary, FAST establishes a highly scalable framework for reinforcement learning and provides a robust solution for the acceleration of large-scale agent training in autonomous driving applications.

\bibliographystyle{IEEEtran}
\bibliography{IEEEexample_new}

\begin{IEEEbiography}
[{\includegraphics[width=1in,height=1.60in, clip,keepaspectratio]{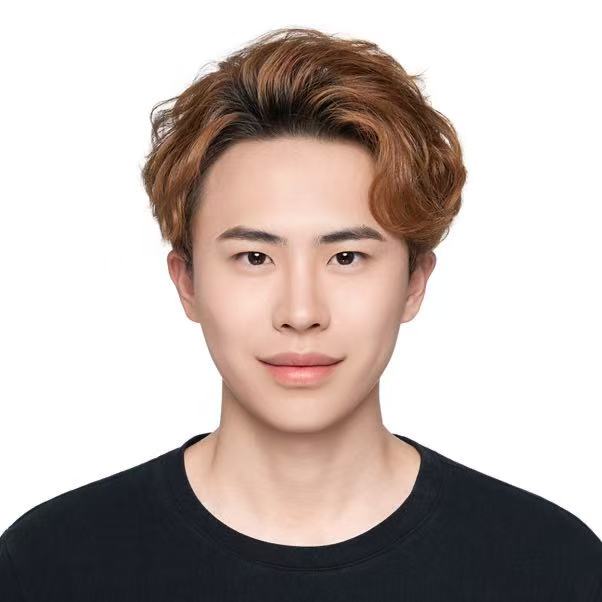}}]
{Bonan Wang} is a research assistant at the School of Vehicle and Mobility, Tsinghua University, Beijing, China. Prior to this, he received his M.S. degree from the State Key Laboratory of Internet of Things for Smart City and the Department of Computer and Information Science at the University of Macau in 2025. He earned his B.S. degree in Data Science and Big Data Technology from Shaanxi University of Science \& Technology in 2023. His research primarily focuses on autonomous driving and reinforcement learning.

\end{IEEEbiography}

\begin{IEEEbiography}
[{\includegraphics[width=1in,height=1.0in, clip,keepaspectratio]{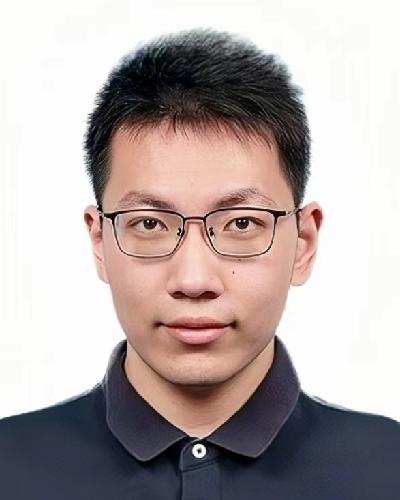}}]
{Letian Tao} received the B.S. degree from the School of Vehicle and Mobility, Tsinghua University, Beijing, China, in 2022, where he is currently pursuing the Ph.D. degree with the School of Vehicle and Mobility, Tsinghua University. His research interests include end-to-end autonomous driving and reinforcement learning.
\end{IEEEbiography}

\begin{IEEEbiography}
[{\includegraphics[width=1in,height=1.0in,clip,keepaspectratio]{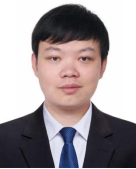}}]
{Bin Shuai} received his Ph.D. in mechanical engineering from the University of Birmingham, Birmingham, UK, in 2021. He is now working as a postdoctoral researcher with Tsinghua-Shuimu Fellowship in the School of Vehicle and Mobility, Tsinghua University, Beijing, China. His research interests include intelligent vehicle control, reinforcement learning, system modeling, and energy management.
\end{IEEEbiography}

\begin{IEEEbiography}
[{\includegraphics[width=1in,height=1.2in,clip,keepaspectratio]{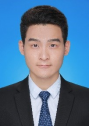}}]
{Jiaxin Gao} received his B.S. and Ph.D. degrees from the University of Science \& Technology Beijing in 2017 and 2023, respectively. He conducted his postdoctoral research at the School of Vehicle and Mobility, Tsinghua University from July 2023 to July 2025. He is currently an Assistant Researcher at the State Key Laboratory of Intelligent Green Vehicle and Mobility, Tsinghua University. His research interests focus on autonomous driving, reinforcement learning, decision and control for autonomous vehicles, perceptual data generation, and simulation data for autonomous driving.
\end{IEEEbiography}

\begin{IEEEbiography}
[{\includegraphics[width=1in,height=1.2in,clip,keepaspectratio]{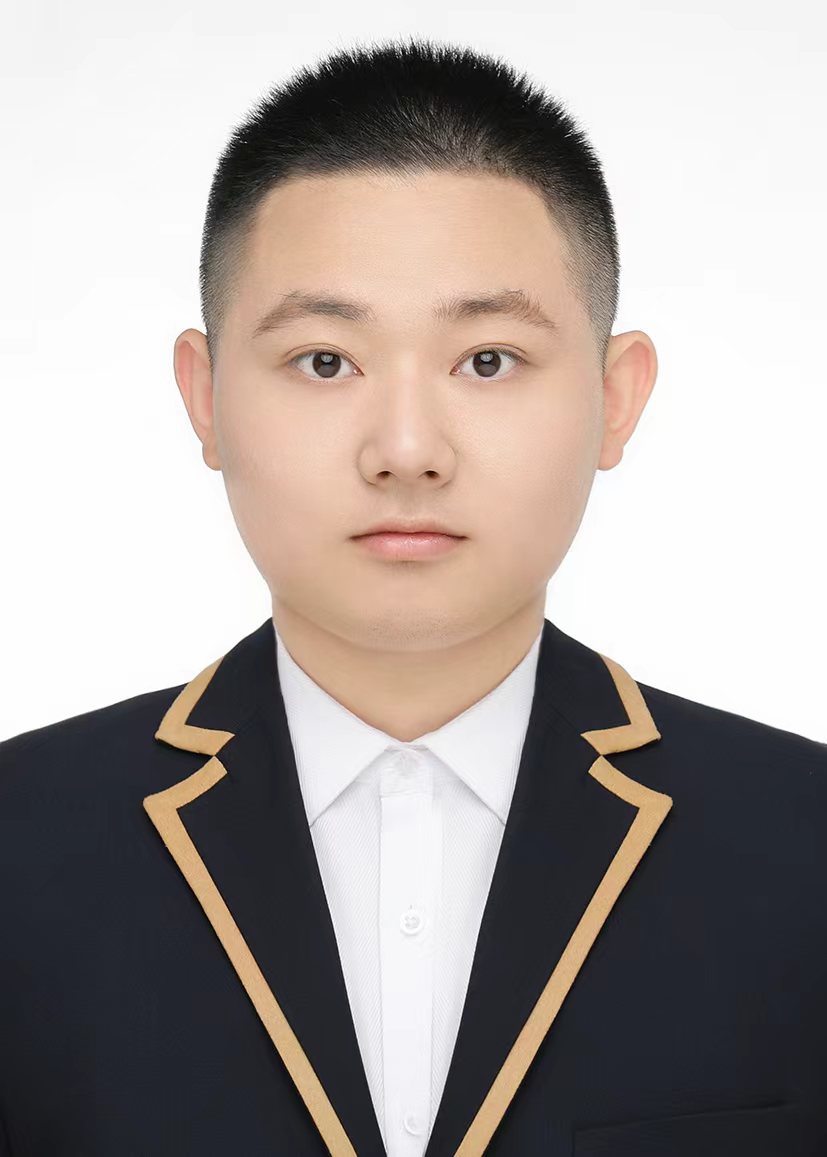}}]
{Wenxin Zhao} received the B.S. degree in Intelligent Science and Technology from Sun Yat-sen University China, in 2025. He is currently a Ph.D. student with the College of AI, Tsinghua University, China. His research interests include Reinforcement Learning, Autonomous Driving and Embodied AI.
\end{IEEEbiography}

\begin{IEEEbiography}
[{\includegraphics[width=1in,height=1.2in,clip,keepaspectratio]{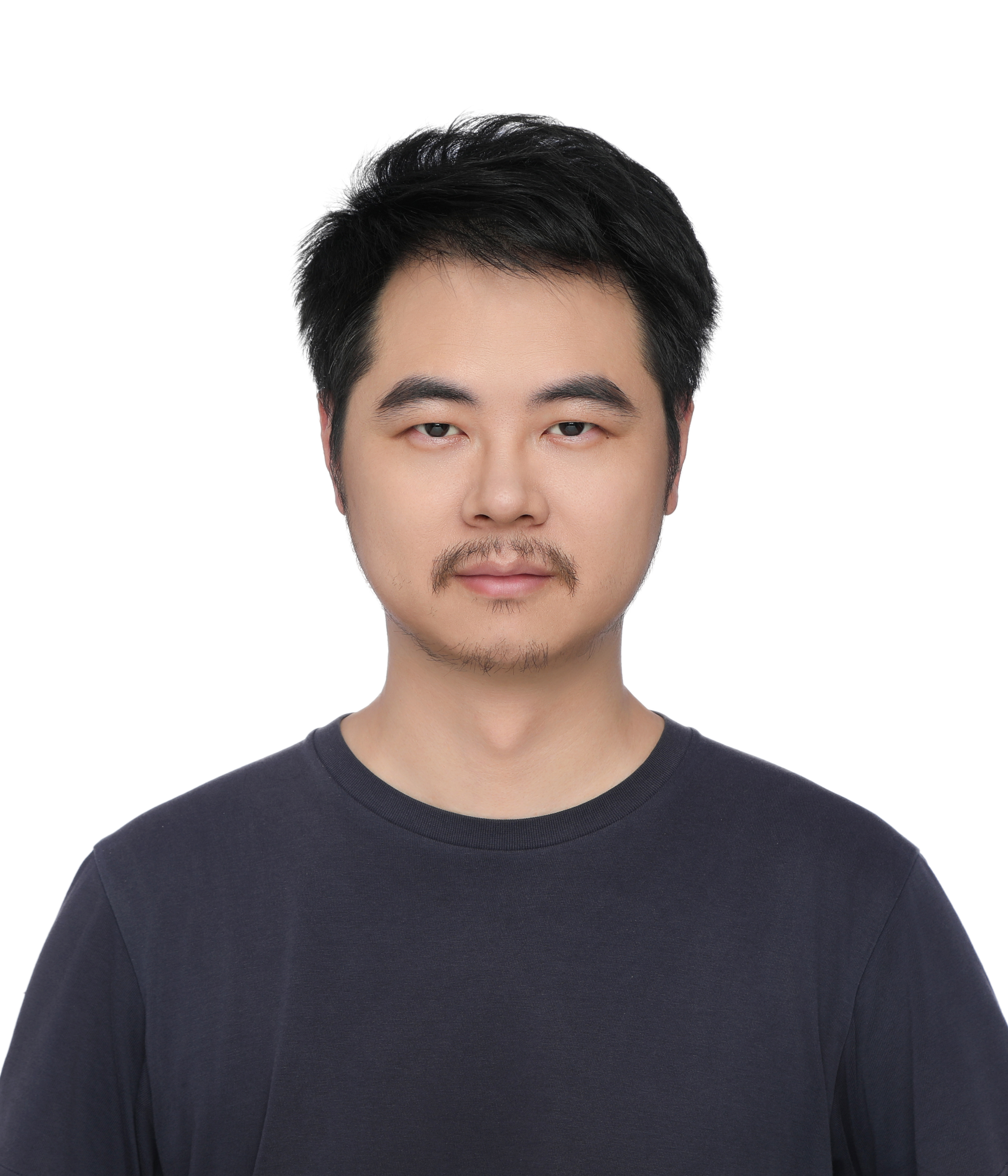}}]
{Wei Xiong}  received his B.Eng. degree from the School of Aerospace Engineering, Tsinghua University in 2009, and his Ph.D. degree in Mechanics from Tsinghua University in 2014. His research interests focus on autonomous driving and artificial intelligence.
\end{IEEEbiography}

\begin{IEEEbiography}
[{\includegraphics[width=1in,height=1.2in,clip,keepaspectratio]{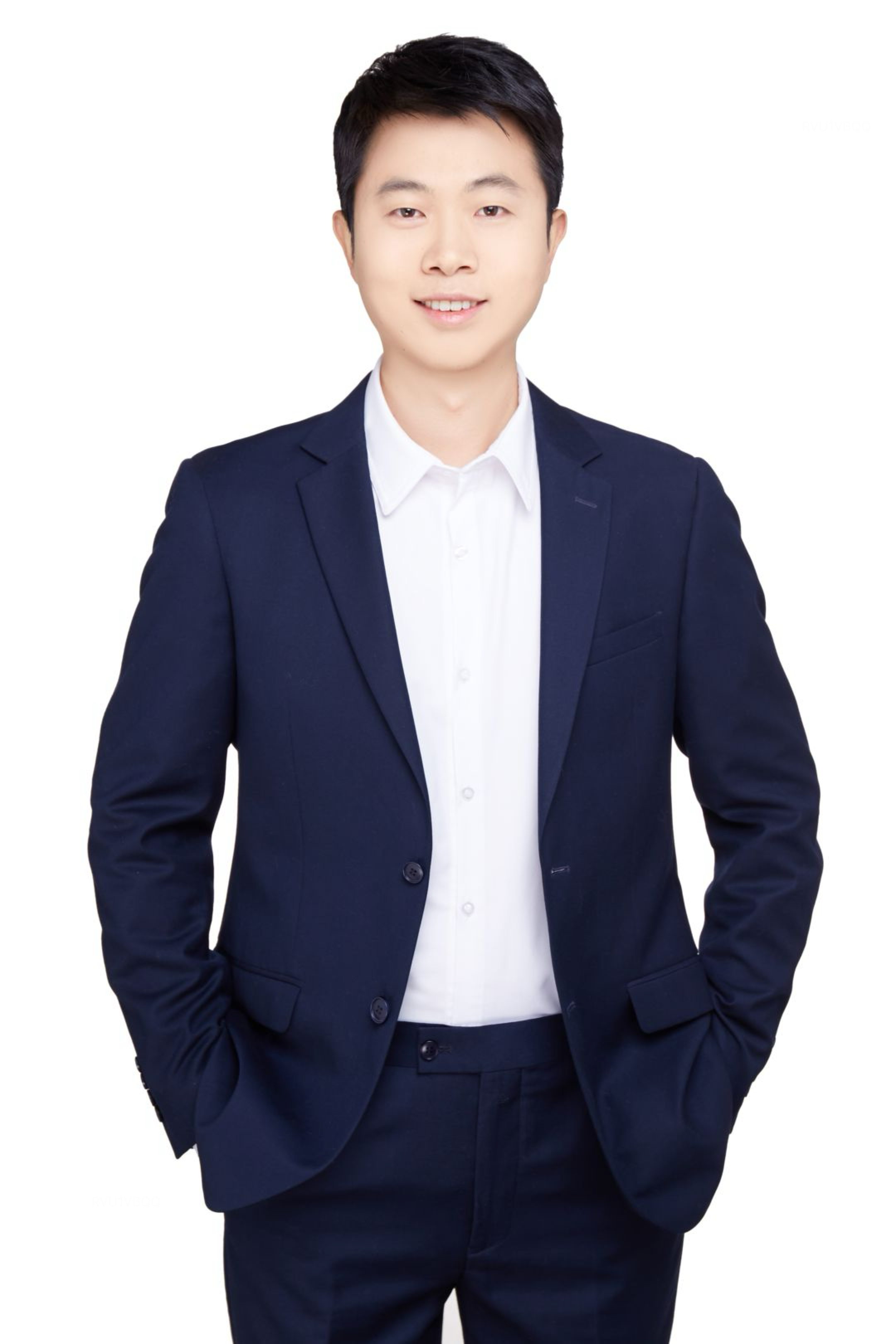}}]
{Kehua Sheng} received his bachelor’s degree from the School of Computer Science and Technology at Hunan University in 2006, and his master’s degree from the School of Computer Science and Technology at Zhejiang University in 2008. His research mainly focuses on human-computer interaction and artificial intelligence.
\end{IEEEbiography}

\begin{IEEEbiography}
[{\includegraphics[width=1in,height=1.2in,clip,keepaspectratio]{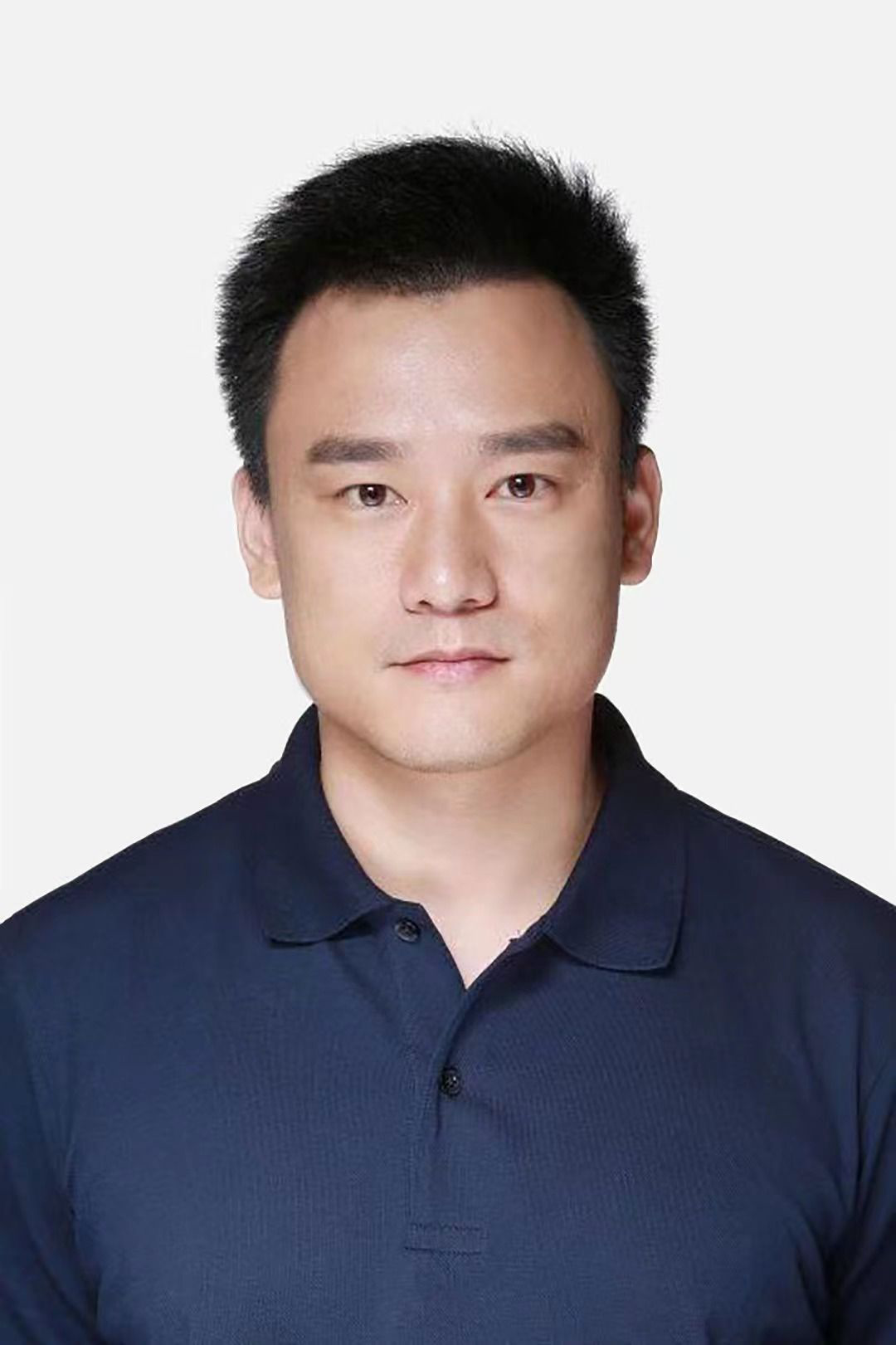}}]
{Bo Zhang} received the B.S. degree from the College of International Software, Wuhan University in 2005, and the M.S. degree from the Institute of Software, Chinese Academy of Sciences in 2009. He is mainly engaged in research on human-computer interaction and artificial intelligence.
\end{IEEEbiography}

\begin{IEEEbiography}
[{\includegraphics[width=1in,height=1.5in,clip,keepaspectratio]{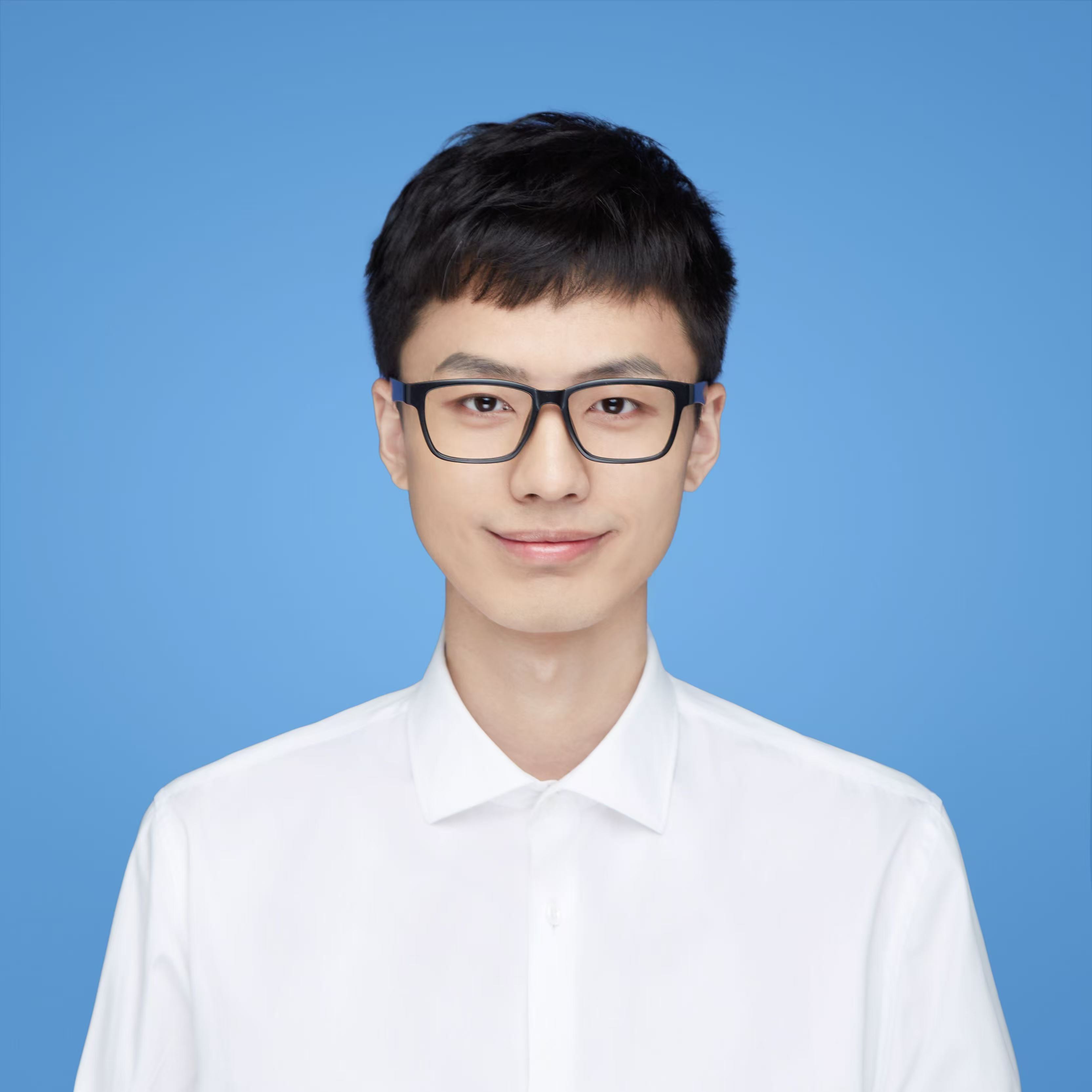}}]
{Yang Guan} received his Ph.D. degree from the School of Vehicle and Mobility, Tsinghua University in 2022. He is currently an assistant researcher in the School of Vehicle and Mobility, Tsinghua University, China. Before joining Tsinghua University, he worked as a research scientist at Meituan Autonomous Driving Department. His active research interests include data-driven planning, end-to-end autonomous driving, reinforcement learning and neural networks. Dr. Guan was the recipient of the Best Student Paper Award at the 23rd and 24th IEEE Intelligent Transportation Systems Conference in 2020 and 2021, respectively, and the IET Premium Award for Best Paper in 2023. He was also a finalist for the Best Student Paper Award at the 32nd IEEE Intelligent Vehicle Symposium in 2021.
\end{IEEEbiography}

\begin{IEEEbiography}
[{\includegraphics[width=1in,height=1.2in,clip,keepaspectratio]{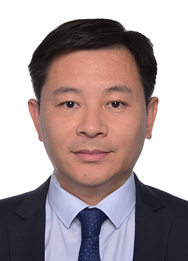}}]
{Shengbo Eben Li} (Senior Member, IEEE) received his M.S. and Ph.D. degrees from Tsinghua University in 2006 and 2009. Before joining Tsinghua University, he has worked at Stanford University, University of Michigan, and UC Berkeley. His active research interests include intelligent vehicles and driver assistance, deep reinforcement learning, optimal control and estimation, etc. He is the author of over 190 peer-reviewed journal/conference papers, and co-inventor of over 40 patents. Dr. Li has received over 20 prestigious awards, including Youth Sci. \& Tech Award of Ministry of Education (annually 10 receivers in China), Natural Science Award of Chinese Association of Automation (First level), National Award for Progress in Sci \& Tech of China, and best (student) paper awards of IET ITS, IEEE ITS, IEEE ICUS, CVCI, etc. He also serves as Board of Governor of IEEE ITS Society, Senior AE of IEEE OJ ITS, and AEs of IEEE ITSM, IEEE TITS, IEEE TIV, IEEE TNNLS, etc.
\end{IEEEbiography}
\vfill

\end{document}